# Solving Multistage Influence Diagrams using Branch-and-Bound Search

**Changhe Yuan, Xiaojian Wu** and **Eric A. Hansen**
Department of Computer Science and Engineering
Mississippi State University
Mississippi State, MS 39762
{cyuan,hansen}@cse.msstate.edu, xw83@msstate.edu

## Abstract

A branch-and-bound approach to solving influence diagrams has been previously proposed in the literature, but appears to have never been implemented and evaluated – apparently due to the difficulties of computing effective bounds for the branch-and-bound search. In this paper, we describe how to efficiently compute effective bounds, and we develop a practical implementation of depth-first branch-and-bound search for influence diagram evaluation that outperforms existing methods for solving influence diagrams with multiple stages.

## 1 Introduction

An influence diagram [7] is a compact representation of the relations among random variables, decisions, and preferences in a domain that provides a framework for decision making under uncertainty. Many algorithms have been developed to solve influence diagrams [2, 3, 8, 15, 17, 18, 19, 21]. Most of these algorithms, whether they build a secondary structure or not, are based on the bottom-up dynamic programming approach. They start by solving small low-level decision problems and gradually build on these results to solve larger problems until the solution to the global-level decision problem is found. The drawback of these methods is that they can waste computation in solving decision scenarios that have zero probability or that are unreachable from any initial state by following an optimal decision policy.

This drawback can be overcome by adopting a branch-and-bound approach to solving an influence diagram that uses a search tree to represent all possible decision scenarios. This approach can use upper bounds on maximum utility to prune branches of the search tree that correspond to low-quality decisions that cannot be part of an optimal policy; it can also prune branches that have zero probability.

A branch-and-bound approach to influence diagram evaluation appears to have been first suggested by Pearl [16]. He proposed it as an improvement over the classic method of unfolding an influence diagram into a decision tree and solving it using the rollback method, which itself is a form of dynamic programming [7]. In Pearl's words:

> A hybrid method of evaluating influence diagrams naturally suggests itself. It is based on the realization that decision trees need not actually be generated and stored in their totality to produce the optimal policy. A decision tree can also be evaluated by *traversing* it in a depth-first, backtracking manner using a meager amount of storage space (proportional to the depth of the tree). Moreover, branch-and-bound techniques can be employed to prune the search space and permit an evaluation without exhaustively traversing the entire tree... an influence diagram can be evaluated by sequentially instantiating the decision and observation nodes (in chronological order) while treating the remaining chance nodes as a Bayesian network that supplies the probabilistic parameters necessary for tree evaluation. (p. 311)

However, neither Pearl nor anyone else appears to have followed up on this suggestion and implemented such an algorithm. The apparent reason is the difficulty of computing effective bounds to prune the search tree. Qi and Poole [17] proposed a similar search-based method for solving influence diagrams, but with no method for computing bounds; in fact, their implementation relied on the trivial *infinity* upper bound to guide the search. Recently, Marinescu [12] proposed a related search-based approach to influence diagram evaluation. But again, he proposed no method for computing bounds; his implementation relies on brute-force search. Even without bounds to prune the search space, note that both Qi and Poole and Marinescu argue that a search-based approach has advantages – for example, it can prune branches that have zero probability.

In this paper, we describe an implemented depth-first branch-and-bound search algorithm for influence diagram evaluation that includes efficient techniques for computing bounds to prune the search tree. To compute effective bounds, our algorithm adapts and integrates two previous contributions. First, we adapt the work of Nilsson and Höhle [14] on computing an upper bound on the maximum expected utility of an influence diagram. The motivation for their work was to bound the quality of strategies found by an approximation algorithm for solving limited-memory influence diagrams, and their bounds are not in a form that can be directly used for branch-and-bound search. We show how to adapt their approach to branch-and-bound search. Second, we adapt the recent work of Yuan and Hansen [20] on solving the MAP problem for Bayesian networks using branch-and-bound search. Their work describes an incremental method for computing upper bounds based on join tree evaluation that we show allows such bounds to be computed efficiently during branch-and-bound search. In addition, we describe some novel methods for constructing the search tree and computing probabilities and bounds that contribute to an efficient implementation. Our experimental results show that this approach leads to an exact algorithm for solving influence diagrams that outperforms existing methods for solving multistage influence diagrams.

## 2 Background

We begin with a brief review of influence diagrams and algorithms for solving them. We also introduce an example of multi-stage decision making that will serve to illustrate the results of the paper.

### 2.1 Influence Diagrams

An influence diagram is a directed acyclic graph $G$ containing variables $\mathbf{V}$ of a decision domain. The variables can be classified into three groups, $\mathbf{V} = \mathbf{X} \cup \mathbf{D} \cup \mathbf{U}$, where $\mathbf{X}$ is the set of oval-shaped *chance* variables that specify the uncertain decision environment, $\mathbf{D}$ is the set of square-shaped *decision* variables that specify the possible decisions to be made in the domain, and $\mathbf{U}$ are the diamond-shaped *utility* variables representing a decision maker's preferences. As in a Bayesian network, each chance variable $X_i \in \mathbf{X}$ is associated with a conditional probability distribution $P(X_i|Pa(X_i))$, where $Pa(X_i)$ is the set of parents of $X_i$ in $G$. Each decision variable $D_j \in \mathbf{D}$ has multiple information states, where an information state is an instantiation of the variables with arcs leading into $D_j$; the selected action is conditioned on the information state. Incoming arcs into a decision variable are called *information arcs*; variables at the origin of these arcs are assumed to be observed before the decision is made. These variables are called the *information variables* of the decision. *No-forgetting* is typically assumed for an influence diagram, which means the information variables of earlier decisions are also information variables of later decisions. We call these past information variables the *history*, and, for convenience, we assume that there are *explicit* information arcs from history information variables to decision variables. Finally, each utility node $U_i \in \mathbf{U}$ represents a function that maps each configuration of its parents to a utility value the represents the preference of the decision maker. (Utility variables typically do not have other variables as children except multi-attribute utility/super-value variables.)

The decision variables in an influence diagram are typically assumed to be temporally ordered, i.e., the decisions have to be made in a particular order. Suppose there are $n$ decision variables $D_1, D_2, ..., D_n$ in an influence diagram. The decision variables partition the variables in $\mathbf{X}$ into a collection of disjoint sets $\mathbf{I_0}, \mathbf{I_1}, ..., \mathbf{I_n}$. For each $k$, where $0 < k < n$, $\mathbf{I_k}$ is the set of chance variables that must be observed between $D_k$ and $D_{k+1}$. $\mathbf{I_0}$ is the set of initial evidence variables that must be observed before $D_1$. $\mathbf{I_n}$ is the set of variables left unobserved when decision $D_n$ is made. Therefore, a partial order $\prec$ is defined on the influence diagram over $\mathbf{X} \cup \mathbf{D}$, as follows:

$$\mathbf{I_0} \prec D_1 \prec \mathbf{I_1} \prec ... \prec D_n \prec \mathbf{I_n}. \qquad (1)$$

A solution to the decision problem defined by an influence diagram is a series of decision rules for the decision variables. A *decision rule* for $D_k$ is a mapping from each configuration of its parents to one of the actions defined by the decision variable. A *decision policy* (or strategy) is a series of decision rules with one decision rule for each decision variable. The goal of solving an influence diagram is to find an *optimal* decision policy that maximizes the expected utility. The maximum expected utility is equal to

$$\sum_{\mathbf{I_0}} \max_{D_1} ... \sum_{\mathbf{I_{n-1}}} \max_{D_n} \sum_{\mathbf{I_n}} P(\mathbf{X}|\mathbf{D}) \sum_j U_j(Pa(U_j)).$$

In general, the summations and maximizations are not commutable. The methods presented in Section 2.3 differ in the various techniques they use to carry out the summations and maximizations in this order.

Recent research has begun to relax the assumption of ordered decisions. In particular, Jensen proposes the framework of unconstrained influence diagrams to allow a partial ordering among the decisions [9]. Other research relaxes the no-forgetting assumption, in particular, the framework of limited-memory influence diagrams [10]. Although the approach we develop can be extended to these frameworks, we do not consider the extension in this paper.

### 2.2 Example

To illustrate decision making using multi-stage influence diagrams, consider a simple maze navigation problem [6,

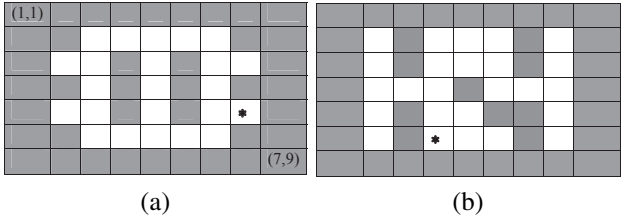

(a)          (b)

Figure 1: Two maze domains. A star represents the goal.

14]. Figure 1 shows four instances of the problem. The shaded tiles represent walls, the white tiles represent movable space, and the white tiles with a star represent goal states. An agent is randomly placed in a non-goal state. It has five available actions that it can use to move toward the goal: it can move a single step in any of the four compass directions, or it can stay in place. The effect of a movement action is stochastic. The agent successfully moves in the intended direction with probability 0.89. It fails to move with probability 0.089, it moves sideways with probability 0.02 (0.01 for each side), and it moves backward with probability 0.001. If movement in some direction would take it into a wall, that movement has probability zero, and the remaining probabilities are normalized. The agent has four sensors, one for each direction, which sense whether the neighboring tile in that direction is a wall. Each sensor is noisy; it detects the presence of a wall correctly with probability 0.9 and mistakenly senses a wall when none is present with probability 0.05. The agent chooses an action at each of a sequence of stages. If the agent is in the goal state after the final stage, it receives a utility value of 1.0; otherwise, it receives a utility value of 0.0.

Figure 2(a) shows the influence diagram for a two-stage version of the maze problem. The variables $x_i$ and $y_i$ represent the location coordinates of the agent at time $i$, the variables $\{ns_i, es_i, ss_i, ws_i\}$ are the sensor readings in four directions at the same time point, and the variable $d_i$ represents the action taken by the agent. The utility variable $u$ assigns a value depending on whether or not the agent is in the goal state after the last action is taken. Since the same variables occur at each stage, we can make the influence diagram arbitrarily large by increasing the number of stages.

### 2.3 Evaluation algorithms

Many approaches have been developed for solving influence diagrams. The simplest is to unfold an influence diagram into an equivalent decision tree and solve it in that form [7]. Another approach called arc reversal solves an influence diagram directly using techniques such as arc-reversal and node-removal [15, 18]; when a decision variable is removed, we obtain the optimal decision rule for the decision. Several methods reduce influence diagrams into Bayesian networks by converting decision nodes into random variables such that the solution of an inference problem in the Bayesian network correspond to the optimal decision policy for the influence diagram [3, 21]. Another method [19] transforms an influence diagram into a valuation network and applies variable elimination to solve the valuation network. Recent work compiles influence diagrams into decision circuits and uses the decision circuits to compute optimal policies [2]; this approach takes advantage of local structure present in an influence diagram, such as deterministic relations.

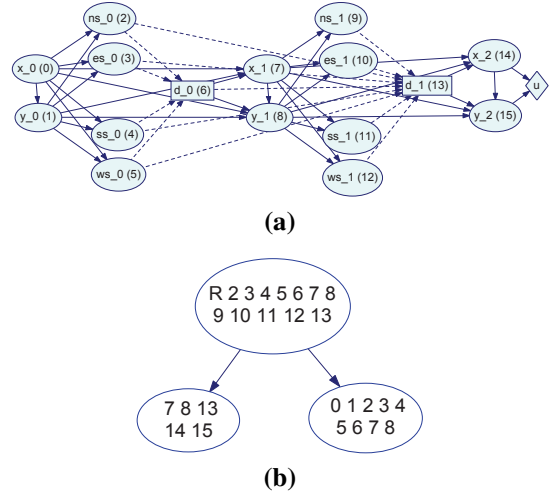

(a)

(b)

Figure 2: (a) An example influence diagram and (b) its strong join tree. The numbers in both figures stand for the indices of the variables. The node with "R" is the strong root.

In the following, we describe a state-of-the-art method for solving an influence diagram using a strong join tree [8]. This method is viewed by many as the fastest general algorithm, and we use its performance as a benchmark for our branch-and-bound approach. A join tree is *strong* if it has at least one clique, $R$, called the *strong root*, such that for any pair of adjacent cliques, $C_1$ and $C_2$, with $C_1$ closer to $R$ than $C_2$, the variables in separator $S = C_1 \cap C_2$ must appear earlier in the partial order defined in Equation (1) than $C_2 \setminus C_1$. A strong join tree for the influence diagram in Figure 2(a) is shown in Figure 2(b).

An influence diagram can be solved exactly by message passing on the strong join tree. Each clique $C$ in the join tree contains two potentials, a probability potential $\phi_C$ and a utility potential $\psi_C$. For clique $C_2$ to send a message to clique $C_1$, $\phi_{C_1}$ and $\psi_{C_1}$ should be updated as follows [8]:

$$\phi'_{C_1} = \phi_{C_1} \times \psi_S; \quad \psi'_{C_1} = \psi_{C_1} + \frac{\psi_S}{\phi_S};$$

where
$$\phi_S = \sum_{C_2 \setminus S} \phi_{C_2}; \quad \psi_S = \max_{C_2 \setminus S} \phi_{C_2} \times \psi_{C_2}.$$

In contrast to the join tree algorithm for Bayesian networks [11], only the collection phase of the algorithm is needed to solve an influence diagram. After the collection phase, we can obtain the maximum expected utility by carrying out the remaining summations and maximizations in the root. In addition, we can extract the optimal decision rules for the decision variables from some of the cliques that contain these variables.

Building a join tree for a Bayesian network may fail if the join tree is too large to fit in memory. This is also true for influence diagrams. In fact, the memory requirement of a strong join tree for an influence diagram is even higher because of the constrained order in Equation (1). Consequently, the join tree algorithm is typically infeasible for solving all but very small influence diagrams.

## 3 Computing bounds

To implement a branch-and-bound algorithm for solving influence diagrams, we need a method for computing bounds – in particular, for computing upper bounds on the utility that can be achieved beginning at a particular stage of the problem, given the history up to that stage. A trivial upper bound is the largest state-dependent value of the utility node of the influence diagram. In this section, we discuss how to compute more informative bounds. There has been little previous work on this topic. Nilsson and Höhle [14] develop an approach to bounding the suboptimality of policies for limited-memory influence diagrams that are found by an approximation algorithm. Dechter [4] describes an approach to computing upper bounds in an influence diagram that is based on mini-bucket partitioning. Neither work considers how to use these bounds in a branch-and-bound search algorithm.

The approach we develop in the rest of this paper is based on the work of Nilsson and Höhle [14], which we extend by showing how it can be used to compute bounds for branch-and-bound search. The general strategy is to create an influence diagram with a value that is guaranteed to be an upper bound on the value of the original influence diagram, but that is also much easier to solve. We use the fact that additional information can only increase the value of an influence diagram. Since this reflects the well-known fact that information value is non-negative, we omit (for space reasons) a proof of the following theorem.

**Theorem 1.** *Let $G$ be an influence diagram and $D$ a decision variable in $G$. Let $\mathbf{I}$ be $D$'s information variables and $\mathbf{R}$ another set of random variables in $G$ that are non-descendants of $D$. Then the influence diagram $G^*$ that results from making $\mathbf{R}$ into information variables by adding information arcs from each variable in $\mathbf{R}$ to $D$ is guaranteed to have a maximum expected utility that is greater than or equal to the maximum expected utility for $G$. We call $G^*$ an* upper-bound influence diagram *for $G$.*

Use of an upper-bound influence diagram to compute bounds only makes sense if the upper-bound influence diagram is simpler and much easier to solve than the original influence diagram. In fact, adding information arcs to an influence diagram can simplify its evaluation by making some other information variables non-requisite. An information variable $I_i$ is said to be *non-requisite* [10, 13] for a decision node $D$ if

$$I_i \perp (\mathbf{U} \cap de(D))|D \cup (Pa(D) \setminus \{I_i\}), \quad (2)$$

where $de(D)$ are the descendants of $D$. A *reduction* of an influence diagram is obtained by deleting all the non-requisite information arcs [14].

Because of the no-forgetting assumption, a decision variable at a late stage may have a large number of history information variables. For decision making under imperfect information, all of these information variables are typically requisite. As a result, the size of the decision rules grows exponentially as the number of decision stages increases, which is the primary reason multi-stage influence diagrams are very difficult to solve.

In constructing an upper-bound influence diagram, we want to add information arcs that make some or all of the history information variables for a decision node non-requisite, in order to simplify the influence diagram and make it easier to solve. We adopt the strategy proposed by Nilsson and Höhle [14]. Let $nd(X)$ be the non-descendant variables of variable $X$, let $fa(X) = Pa(X) \cup \{X\}$ be the family of variable $X$ (i.e., the variable and its parents), let $fa(\mathbf{X}) = \cup_{X_i \in \mathbf{X}} fa(X_i)$ be the family of the set of variables $\mathbf{X}$ (i.e., the variables and their parents), and let $\triangle_j$ be $\{D_1, ...D_j\}$ be a sequence of decision variables from stage 1 to stage $j$. The following theorem is proved by Nilsson and Höhle [14].

**Theorem 2.** *For an influence diagram with the constrained order in Equation (1), if we add to each decision variable $D_j$ the following new information variables in the order of $j = n, ..., 1$,*

$$N_j = \arg\min_{B \subseteq (B_j \cap nd(D_j))}\{|B| | fa(\triangle_j) \\ \perp (\mathbf{U} \cap de(D_j))|(B \cup \{D_j\})\}, \quad (3)$$

*where*

$$B_j = \begin{cases} \mathbf{U} \cup \mathbf{D} & j=k, \\ \cap_{i=j+1}^{k}\{n \in V | n \perp (\mathbf{U} \cap de(D_j))|fa(D_i)\} & j<k, \end{cases}$$

*the following holds for any $D_j$ in the resulting influence diagram:*

$$(de(D_j) \cap \mathbf{U}) \perp fa(\triangle_{i-1})|fa(D_j). \quad (4)$$

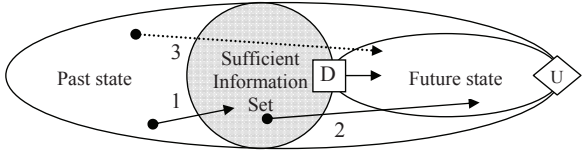

Figure 3: Relations between past and future information states and the minimum sufficient information set.

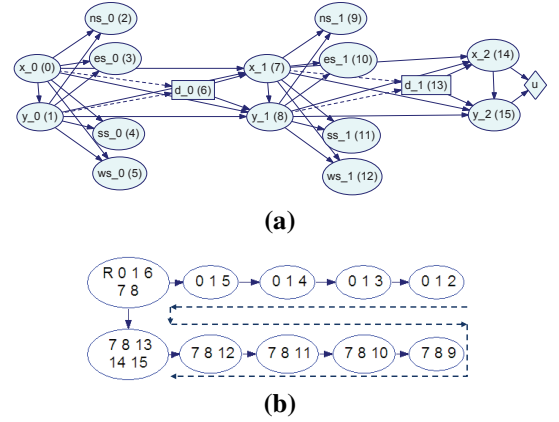

**(a)**

**(b)**

Figure 4: (a) the upper-bound influence diagram for the diagram in Figure 2(a), and (b) its strong join tree.

What this theorem means is that for each decision variable $D_j$ in an influence diagram, there is a set of information variables $N_j$ such that the optimal policy for $D_j$ depends only on these information variables, and is otherwise history-independent. Note that the set $N_j$ for decision variable $D_j$ can contain both information variables from the original influence diagram and information variables created by adding new information arcs to the diagram. The set $N_j$ of information variables can be interpreted as the current "state" of the decision problem, such that if the decision maker knows the current state, it does not need to know any previous history in order to select an optimal action; in this sense, the state satisfies the Markov property.

Consider a decision-making problem in a partially observable domain, such as the maze domain of Section 2.2. The agent cannot directly observe its location in the maze and must rely on imperfect sensor readings to infer its location. In this domain, adding information arcs from the location variables to a decision variable, so that the agent know the current location at the time it chooses an action, makes both current and history sensor readings non-requisite, which results in a much simpler influence diagram, which in this case serves as an upper-bound influence diagram.

We call $N_j$ a *sufficient information set* (SIS) for $D_j$. The intuition behind this approach is illustrated by Figure 3. The shaded oval shows the SIS set $N_D$ for decision $D$. The past state affects the variables in $N_D \cup \{D\}$, illustrated by the arc labeled 1, and $N_D \cup \{D\}$ affects the future state, as illustrated by arc 2. The future state further determines the values of the utility variables. $N_D \cup \{D\}$ d-separates the past and future states and prevents the direct influence shown by arc 3. The concept of a sufficient information set is related to the concept of *extremality*, as defined in [21], and the concept of *blocking*, as defined in [13].

To construct an upper-bound influence diagram, we find the SIS set for each decision in the order of $D_n, ..., D_1$ and make the variables in each SIS set information variables for the corresponding decisions. We then delete the non-requisite information arcs. Consider the influence diagram in Figure 4(a) as an example. The information set for $d_1$ is originally $\{ns_0, es_0, ss_0, ws_0, d_0, ns_1, es_1, ss_1, ws_1\}$. We find that its sufficient information (SIS) set is $\{x_1, y_1\}$. We also find that the SIS set for $d_0$ is $\{x_0, y_0\}$. By making $\{x_1, y_1\}$ and $\{x_0, y_0\}$ information variables for $d_1$ and $d_0$ respectively, and reducing the influence diagram, we obtain the much simpler influence diagram in Figure 4(a). The strong join tree for the new influence diagram is shown in Figure 4(b), which is also much smaller than the strong join tree for the original model. Since the upper-bound influence diagram assumes the actual location is directly observable to the agent, it effectively transforms a partially observable decision problem into a fully observable one. The resulting influence diagram and, hence, its join tree is much easier to solve.

Finding the sufficient information set (SIS) for a decision variable in an influence diagram is equivalent to finding a minimum separating set in the moralized graph of the influence diagram [14]. Acid and de Campos [1] propose an algorithm based on the *Max-flow Min-cut* algorithm [5] for finding a minimum separating set between two sets of nodes in a Bayesian network with some of the separating variables being fixed. We use their algorithm to find the SIS sets. The two sets of nodes are $fa(\triangle_j)$ and $\mathbf{U} \cap de(D_j)$. The only fixed separating variable is $D_j$. The algorithm first introduces two dummy variables, *source* and *sink*, to the moralized graph. The source is connected to the neighboring variables of $fa(\triangle_j)$, and the sink to the variables in $de(D_j) \cap an(\mathbf{U} \cap de(D_j))$. We then create a max-flow network out of the undirected graph by assigning each edge capacity 1.0. A solution gives a minimum separating set between the sink and source that contains $D_j$.

We briefly mention some issues that are not described by Nilsson and Höhle [14], but that need to be considered in an implementation. The first issue is how to define the size of an SIS set. Theorem 2 uses the *cardinality* of the SIS set as the minimization criterion. Another viable choice is to use *weight*, defined as the product of number of states, of the variables in an SIS set as the minimization criterion.

The relation between these two criteria is similar to the relation between *treewidth* and weight in constructing a junction tree. While treewidth tells us how complex a Bayesian network is at the structure level, weight provides an idea on how large the potentials of the junction tree are at the quantitative level. Both methods have been used. In our implementation, we use the cardinality.

A second issue is that multiple candidate SIS sets may exist for a decision variable. In that case, we need some criterion for selecting the best one. In our implementation, we select the candidate SIS set that is *closest* to the descendant utility variables of the decision. Note that other candidate sets are all d-separated from the utility node by the closest SIS set. This choice has the advantage that the resulting influence diagram is easiest to evaluate; however, other choices may result in an influence diagram that gives a tighter bound.

## 4 Branch-and-bound search

In this section, we describe how to use the upper-bound influence diagram to compute bounds for a depth-first branch-and-bound search algorithm that solves the original influence diagram. We begin by showing how to represent the search space as an AND/OR tree. A naive approach to computing bounds requires evaluating the entire upper-bound influence diagram at each OR node of the search tree, which is computationally prohibitive. To make branch-and-bound search feasible, we rely on an incremental approach to computing bounds proposed by Yuan and Hansen [20] for solving the MAP problem in Bayesian networks using branch-and-bound search. We show how to adapt that approach in order to solve influence diagrams efficiently.

### 4.1 AND/OR tree search

We represent the search space for influence diagram evaluation as an AND/OR tree. The nodes in an AND/OR tree are of two types: AND nodes and OR nodes. AND nodes correspond to chance variables; a probability is associated with each arc originating from an AND node and the probabilities of all the arcs from an AND node sum to $1.0$. The OR nodes correspond to decision variables. Each of the leaf nodes of the tree has a utility value that is derived from the utility node of the influence diagram.

Qi and Poole [17] create an AND/OR tree in which each layer of AND nodes alternates with a layer of OR nodes. Each AND node in this tree corresponds to the information set of a decision variable in the influence diagram, which is a set of information variables. To compute the probability for each arc emanating from an AND node in this tree, however, it is necessary to have the joint probability distributions of all the information sets; these are often not readily available, since variables in the same informa-tion set can belong to different clusters of a join tree. For computational convenience, our AND/OR tree is based on the structure of the strong join tree in Figure 4(b). For the maze example, we order the variables in the information set $\{ns_0, es_0, ss_0, ws_0\}$ according to the order in Equation (5). Note that the join tree does not have a clique that contains all four variables; in fact, they are all in different cliques. So we consider the variables one by one. That means that our AND/OR tree allows several layers of AND nodes to alternate with a layer of OR nodes. See Figure 5 for an example of the kind of AND/OR tree constructed by our search algorithm, and note that the first four layers of this AND/OR tree are all AND layers.

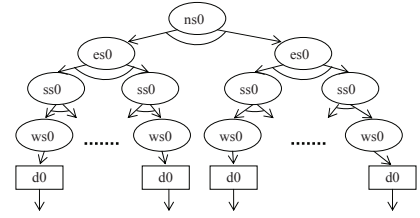

Figure 5: The AND/OR tree used in our approach. Oval-shaped nodes are AND nodes, and square-shaped nodes are OR nodes.

Each path from the root of the AND/OR tree to a leaf corresponds to a complete instantiation of the information variables and decision variables of the influence diagram, that is, a complete history. Since we use the AND/OR tree to find an optimal decision policy for the original influence diagram, we have to construct an AND/OR tree that is consistent with the original constrained order. For the influence diagram in Figure 2(a), the partial order is

$$\{ns_0, es_0, ss_0, ws_0\} \prec \{d_0\} \prec \{ns_1, es_1, ss_1, ws_1\}$$
$$\prec \{d_1\} \prec \{x_0, y_0, x_1, y_1, x_2, y_2, u\}, \qquad (5)$$

and the decision variables must occur in this order along any branch.

We define a *valuation function* for each node in an AND/OR tree as follows: (a) for a leaf node, the value is its utility value; (b) for an AND node, the value of is the sum of the values of its child nodes weighted by the probabilities of the outgoing arcs; (c) for an OR node, the value is the maximum of the summed utility values of each child node and corresponding arc. We use this valuation function to find an optimal strategy for the influence diagram.

We represent a strategy for a multi-stage influence diagram as a policy tree that is defined as follows. A *policy tree* of an AND/OR tree is a subtree such that: (a) it consists of the root of the AND/OR tree; (b) if a non-terminal AND node is in the policy tree, all its children are in the policy tree; and (c) if a non-terminal OR node is in the policy tree, exactly one of its children is in the policy tree. Given an AND/OR tree that represents all possible histories and strategies for an influence diagram, the influence diagram is solved by finding a policy tree with the maximum value at the root, where the value of the policy tree is computed based on the valuation function. Depth-first branch-and-bound search can be used to find an optimal policy tree.

The AND/OR tree is constructed on-the-fly during the branch-and-bound search, and it is important to do so in a way that allows the probabilities and values to be computed as efficiently as possible. We use the maze problem and the AND/OR tree in Figure 5 as an example. (The upper-bound influence diagram and join tree are shown in Figure 4.) We have already pointed out that including more layers in our AND/OR tree allows us to more easily use the probabilities and utilities computed by the join tree. If we start by expanding $ns_0$ (where expanding a node refers to generating its child nodes in the AND/OR tree), we need the probabilities of $P(ns_0)$ to label the outgoing arcs. We can readily look up the probabilities from clique $(0, 1, 2)$ after an initial full join tree evaluation. Note that we use the join tree of the upper-bound influence diagram to compute the probabilities. We can do that because these probabilities are the same as those computed from the original influence diagram. This is due to the fact that the same set of actions will reduce both models into the same Bayesian networks. Adding information arcs to an influence diagram, in order to create an upper-bound influence diagram, only changes the expected utilities of the decision variables.

After expanding $ns_0$, we expand any of $\{es_0, ss_0, ws_0\}$. Suppose the next variable is $es_0$, we need the conditional probabilities of $es_0$ given $ns_0$. These probabilities can be computed by setting the state of $ns_0$ as new evidence to the join tree and evaluating the join tree again. The same process is used in expanding $\{ss_0, ws_0\}$.

Note that we do not have to expand one variable at a time. If a clique has multiple variables in the same information set, the variables can be expanded *together* because a joint probability distribution over them can be easily computed. Expanding them together also saves the need to do marginalization. For example, variables $x_1, y_1, x_2, y_2$ (7,8,14,15) are in the same information group and also reside in a same clique. In this case, we can expand them as a single layer in the AND/OR tree.

After $\{ns_0, es_0, ss_0, ws_0\}$ are expanded, we expand $d_0$ as an OR layer. This is where the upper bounds are needed. We set the states of $\{ns_0, es_0, ss_0, ws_0\}$ as evidence to the join tree and compute the expected utility values for $d_0$ by reevaluating the join tree. The expected utilities of $d_0$ are upper bounds for the same decision scenarios of the original model. We can use the upper bounds to select the most promising decision alternative to search first. The exact value will be returned once the subtree is searched. If the value is higher than the upper bounds of the remaining decision alternatives, these alternatives are immediately pruned because they cannot be part of an optimal decision policy. We repeat the above process until a complete policy tree is found.

### 4.2 Incremental join tree evaluation

It is clear that repeated join tree evaluation has to be performed in computing the upper bounds and conditional probabilities. A naive approach is at each time to set the states of instantiated variables as evidence and perform a *full* join tree evaluation. However, that is too costly. We can use an efficient *incremental* join tree evaluation method to compute the probabilities and upper-bound utilities, based on the incremental join tree evaluation method proposed by Yuan and Hansen [20] for solving the MAP problem.

The key idea is that we can choose a particular order of the variables that satisfies the constraints of Equation (5) such that an *incremental join tree evaluation scheme* can be used to compute the information. Given such an order, we know exactly which variables have been searched and which variable will be searched next at each search step. We only need to send messages from the parts of the join tree that contain the already searched variables to a clique with the next search variable. For example, after we search $es_0$, the only message needs to be sent to obtain $P(ns_0|es_0)$ is the message from clique $(0, 1, 3)$ to $(0, 1, 2)$. There is no need to evaluate the whole join tree. If we choose the following search order for the maze problem

$$ns_0, es_0, ss_0, ws_0, d_0, ns_1, es_1, ss_1, ws_1, d_1, x_0, y_0,$$
$$x_1, y_1, x_2, y_2,$$

we can use an incremental message passing in the direction of the dashed arc in Figure 4(b) to compute the probabilities and upper bounds needed in one downward pass of a depth-first search.

Both message collection and distribution are needed in this new scheme, unlike evaluating a strong join tree for the original influence diagram. The messages being propagated contain two parts: utility potentials and probability potentials. The distribution phase is typically needed to compute the conditional probabilities. For example, suppose we decide to expand $es_0$ before $ns_0$, we have to send a message from clique $(0, 1, 3)$ to $(0, 1, 2)$ to obtain $P(ns_0|es_0)$. We only need to send the probability potential in message distribution. We do not need to send utility potentials because past payoffs do not count towards the expected utilities of future decisions.

### 4.3 Efficient backtracking

We use *depth-first branch-and-bound* (DFBnB) to utilize the efficient incremental bound computation. Depth-first search makes sure that the search need not jump to a different search branch before backtracking is needed. In other words, the join tree only needs to work with one search history at a time.

|   | stages | Join tree | | DFBnB | | | | | Exhaustive search | | |
|---|---|---|---|---|---|---|---|---|---|---|---|
|   |   | time | mem. | time | mem. | policy | #bounds | #zeros | time | mem. | #zeros |
| a | 2 | 15 | 8.0 | 125 | 3.2 | 783 | 816 | 0 | 703 | 3.1 | 0 |
|   | 3 | 640 | 238.4 | 1s812 | 4.7 | 12,559 | 13,104 | 0 | 43s218 | 4.7 | 0 |
|   | 4 | - | - | 59s610 | 23.9 | 200,975 | 446,255 | 0 | 47m42s625 | 25.1 | 0 |
|   | 5 | - | - | 32m55s766 | 343.0 | 3,215,631 | 14,546,815 | 0 | - | - | - |
| b | 2 | 16 | 8.0 | 109 | 3.2 | 783 | 816 | 0 | 688 | 3.1 | 0 |
|   | 3 | 640 | 238.4 | 2s109 | 4.7 | 12,559 | 14,734 | 0 | 43s953 | 4.7 | 0 |
|   | 4 | - | - | 43s641 | 23.9 | 200,975 | 325,820 | 0 | 49m00s516 | 25.1 | 0 |
|   | 5 | - | - | 18m00s437 | 323.1 | 3,215,631 | 7,883,235 | 0 | - | - | - |

Table 1: Comparison of three algorithms (join tree algorithm, DFBnB using the join tree bounds, and exhaustive search of the AND/OR tree) in solving maze problems *a* and *b* for different numbers of stages. The symbol '-' indicates the problem could not be solved, due to memory limits in the case of the join tree algorithm, or due to a three-hour time limit in the case of the search algorithms. Running time is in hours (h), minutes (m), seconds (s) and milliseconds. Memory (mem.) is in megabytes and is the peak amount of memory used by the algorithm; "policy" is the count of nodes (both OR and AND nodes) in the part of the search tree containing the optimal policy tree – it is the same for both DFBnB and exhaustive search; "#bounds" is the count of times a branch from an OR node was pruned based on bounds; "#zeros" is the count of times a branch from an AND node was pruned because it had zero probability.

We do need to backtrack to a previous search node once we finish a search branch or realize that a search branch is not promising and should be pruned. We need to retract all the newly set evidence since the generation of that search node and restore the join tree to a previous state. One way to achieve this is to reinitialize the join tree with correct evidence and perform a full join tree evaluation, which we have already pointed out is too costly. Instead, we cache the potentials and separators along the message propagation path before changing them by either setting evidence or overriding them with new messages. When backtracking, we simply restore the most recently cached potentials and separators in the reverse order. The join tree will be restored to the previous state with no additional computations. This backtracking method is much more efficient than reevaluating the whole join tree. For solving the MAP problem for Bayesian networks, Yuan and Hansen [20] show that the incremental method is more than ten times faster than full join tree evaluation in depth-first branch-and-bound search.

## 5 Empirical Evaluation

We compared the performance of our algorithm against both the join tree algorithm [8] and exhaustive search of the AND/OR tree (i.e., no bounds are used for pruning). Experiments were run on a 2.4 GHz Duo processor with 3 gigabytes of RAM running Windows XP.

The results in Table 1 are for the two maze problems in Figure 1, which we solved for different numbers of stages. When there are only two or three stages, the join tree algorithm is most efficient. This is because the strong join trees for these influence diagrams are rather small and can be built successfully. Once the join trees are built, only one collection phase is necessary to solve the influence diagram; by contrast, the depth-first branch-and-bound algorithm (DFBnB) algorithm must perform repeated message propagations to compute upper bounds and probabilities during the search. For more then three stages, however, the join tree algorithm cannot solve the maze models because the strong join trees are too large to fit in memory. Because the exhaustive search algorithm only needs to store the search tree and policy, it can solve the maze models for up to four stages, although it takes more then 45 minutes to do so. The DFBnB algorithm can solve the maze models for up to five stages in less time than it takes the exhaustive search algorithm to solve them for four stages, demonstrating the advantage of using bounds to prune the search tree. Table 1 includes the number of times a branch of the search tree is pruned based on bounds, as well as the number of times a branch with zero probability is pruned. For the maze models with their original parameter settings, every branch of the search tree has non-zero probability.

Previous work has argued that one of the advantages of search algorithms for influence diagram evaluation is that they can prune branches of the search tree that have zero probability, even without bounds [12, 17]. To test this argument, as well as to illustrate the effect of different problem characteristics on algorithm performance, we modified the maze models described in Section 2.2. First, we removed some noise from the sensors. Each of the four sensors reports a wall in the corresponding direction of the compass. In the original problem, each sensor is noisy; it detects the presence of a wall correctly with probability 0.9 and mistakenly senses a wall when none is present with probability 0.05. As a result, every sensor reading is possible in every state and there are no zero-probability branches. We modified the model so that each sensor accurately detects whether a wall is present in its direction of the compass. With this change, the maze problem remains partially observable, but the search tree contains many zero-probability branches, as can be seen from the results in Table 2. Since the search algorithms can prune zero-probability branches, the exhaustive search algorithm can now solve the problem for up to five stages and the DFBnB algorithm can solve the problem for up to six stages.

| Maze domains modified by removing some noise from sensors | | | | | | | | | | | |
|---|---|---|---|---|---|---|---|---|---|---|---|
| | | Join tree | | DFBnB | | | | | Exhaustive search | | |
| | stages | time | mem. | time | mem. | policy | #bounds | #zeros | time | mem. | #zeros |
| a | 2 | 16 | 8.0 | 140 | 3.1 | 224 | 34 | 246 | 62 | 3.1 | 312 |
| | 3 | 641 | 238.4 | 1s172 | 3.6 | 849 | 107 | 2,598 | 781 | 3.5 | 4,616 |
| | 4 | - | - | 15s281 | 4.3 | 2,843 | 798 | 28,282 | 9s828 | 4.1 | 61,320 |
| | 5 | - | - | 2m25s266 | 5.5 | 9,076 | 6,282 | 325,146 | 4m04s062 | 4.3 | 773,768 |
| | 6 | - | - | 12m03s828 | 6.3 | 28,413 | 41,921 | 2,885,925 | 1h0m26s078 | 7.9 | 9,652,872 |
| b | 2 | 15 | 8.0 | 109 | 3.2 | 261 | 51 | 175 | 79 | 3.1 | 292 |
| | 3 | 641 | 238.4 | 1s203 | 3.6 | 1,136 | 186 | 2,507 | 1s109 | 3.5 | 5,876 |
| | 4 | - | - | 14s203 | 4.4 | 4,244 | 660 | 29,458 | 16s391 | 4.3 | 88,948 |
| | 5 | - | - | 1m39s046 | 5.9 | 15,030 | 2,302 | 229,003 | 7m59s218 | 6.0 | 1,255,284 |
| | 6 | - | - | 5m14s047 | 10.0 | 52,240 | 58,058 | 1,442,212 | 1h57m22s625 | 10.9 | 17,418,100 |

| Maze domains modified by removing some noise from both actions and sensors | | | | | | | | | | | |
|---|---|---|---|---|---|---|---|---|---|---|---|
| | | Join tree | | DFBnB | | | | | Exhaustive search | | |
| | stages | time | mem. | time | mem. | policy | #bounds | #zeros | time | mem. | #zeros |
| a | 2 | 16 | 8.0 | 110 | 3.2 | 177 | 30 | 206 | 109 | 3.1 | 276 |
| | 3 | 594 | 238.4 | 750 | 3.7 | 562 | 99 | 1,818 | 1s015 | 3.5 | 3,192 |
| | 4 | - | - | 5s359 | 4.1 | 1,504 | 276 | 13,276 | 10s234 | 4.0 | 33,660 |
| | 5 | - | - | 50s453 | 4.8 | 4,216 | 807 | 124,312 | 1m44s406 | 4.7 | 343,843 |
| | 6 | - | - | 5m09s328 | 6.0 | 11,198 | 2,236 | 767,084 | 17m39s812 | 5.9 | 3,490,703 |
| | 7 | - | - | 36m59s906 | 8.3 | 29,653 | 5,947 | 4,821,057 | - | - | - |
| b | 2 | 15 | 8.0 | 94 | 3.1 | 203 | 45 | 164 | 156 | 3.1 | 278 |
| | 3 | 500 | 238.4 | 672 | 3.7 | 727 | 165 | 1,602 | 1s391 | 3.5 | 4,047 |
| | 4 | - | - | 5s297 | 4.1 | 2,029 | 524 | 13,601 | 14s984 | 4.0 | 46,812 |
| | 5 | - | - | 27s688 | 5.1 | 5,540 | 1,557 | 72,518 | 2m36s313 | 4.9 | 504,264 |
| | 6 | - | - | 2m04s812 | 6.7 | 15,625 | 4,941 | 335,185 | 30m24s453 | 6.6 | 5,312,539 |
| | 7 | - | - | 18m13s157 | 10.5 | 43,673 | 16,639 | 2,984,966 | - | - | - |

Table 2: Comparison of three algorithms (join tree algorithm, DFBnB using the join tree bounds, and exhaustive search of the AND/OR tree) in solving maze problems *a* and *b* with modified parameters; for the results in the top table, some noise is removed from the sensors only; for the results in the bottom table, some noise is removed from the actions and the sensors. Removing some noise from the actions and sensors results in more zero-probability branches that can be pruned, allowing the search algorithms (but not the join tree algorithm) to solve the problem for a larger number of stages.

We next made an additional change to the transition probabilities for actions. In the original problem, the agent successfully moves in the intended direction with probability 0.89 (as long as there is not a wall). It fails to move with probability 0.089, it moves sideways with probability 0.02 (0.01 for each side), and it moves backward with probability 0.001. We modified these transition probabilities so that the agent still moves in the intended direction with probability 0.89; but otherwise, it stays in the same position with probability 0.11. The effects of the agent's actions are still stochastic, but they are more predictable, and this allows the search tree to be pruned even further. As a result, the exhaustive search algorithm can solve the problem for up to six stages and the DFBnB algorithm can solve the problem for up to seven stages.

Note that changing the problem characteristics has no effect on the performance of the join tree algorithm. The join tree algorithm solves the influence diagram for all information states, including those that have zero probability and those that are unreachable from the initial state; as a result, its memory requirements explode exponentially in the number of stages and the algorithm quickly becomes infeasible. Although the policy tree that is returned by the search algorithms can also grow exponentially in the number of stages, it does so much more slowly because so many branches can be pruned. As is vividly shown by the results for the two different mazes and for different parameter settings, the performance of the search algorithms is sensitive to problem characteristics – precisely because the search algorithms exploit a form of problem structure that is not exploited by the join tree algorithm. In addition, the results show the effectiveness of bounds in scaling up the search-based approach.

# 6 Conclusion

We have described the first implementation of a depth-first branch-and-bound search algorithm for influence diagram evaluation. Although the idea has been proposed before, we adapted and integrated contributions from related work and introduced a number of new ideas to make the approach computationally feasible. In particular, we described an efficient approach for using the join tree algorithm to compute upper bounds to prune the search tree. The idea is to generate an upper-bound influence diagram by allowing each decision variable to be conditioned on additional information that makes the remaining history non-requisite, thus simplifying the influence diagram. Then a join tree of the upper-bound influence diagram is used to incrementally compute upper bounds for the depth-first branch-and-bound search. We have also described a new approach to constructing the search tree based on the structure of the strong join tree of the upper-bound influence diagram. Experiments show that the resulting depth-first branch-and-bound search algorithm outperforms the state-of-the-art join tree algorithm in solving multistage influence diagrams, at least when there are more than three stages.

We are currently considering how to extend this approach to solve limited-memory influence diagrams [10], which typically have many more stages. We are also exploring approaches to compute more accurate bounds for pruning the search tree. Finally, we are considering approximate and bounded-optimal search algorithms for solving larger influence diagrams using the same upper bounds and AND/OR search tree.

**Acknowledgments** This research was support in part by NSF grants IIS-0953723 and EPS-0903787, and by the Mississippi Space Grant Consortium and NASA EPSCoR program.

# References

[1] S. Acid and L. de Campos. An algorithm for finding minimum d-separating sets in belief networks. In *Proceedings of the 12th Annual Conference on Uncertainty in Artificial Intelligence (UAI-96)*, pages 3–10, San Francisco, CA, 1996. Morgan Kaufmann.

[2] D. Bhattacharjya and R. Shachter. Evaluating influence diagrams with decision circuits. In *Proceedings of the Twenty-Third Conference on Uncertainty in Artificial Intelligence*, pages 9–16. AUAI Press, 2007.

[3] G. Cooper. A method for using belief networks as influence diagrams. In *Proceedings of the Fourth Conference on Uncertainty in Artificial Intelligence*, pages 55–63, Minneapolis, MN, USA, 1988. Elsevier Science, New York, NY.

[4] R. Dechter. An anytime approximation for optimizing policies under uncertainty. In *Workshop on Decision Theoretic Planning, at the 5th International Conference on Artificial Intelligence Planning Systems (AIPS-2000)*, Breckenridge, CO, USA, 2000.

[5] L. R. Ford and D. R. Fulkerson. Maximal flow through a network. *Canadian Journal of Mathematics*, 8:399–404, 1956.

[6] M. C. Horsch and D. Poole. An anytime algorithm for decision making under uncertainty. In *Proceedings of the 14th Annual Conference on Uncertainty in Artificial Intelligence (UAI-98)*, 1998.

[7] R. A. Howard and J. E. Matheson. Influence diagrams. In R. A. Howard and J. E. Matheson., editors, *The Principles and Applications of Decision Analysis*, pages 719–762, Menlo Park, CA, 1981.

[8] F. Jensen, F. V. Jensen, and S. L. Dittmer. From influence diagrams to junction trees. In *Proceedings of the Tenth Conference on Uncertainty in Artificial Intelligence*, pages 367–373. Morgan Kaufmann, 1994.

[9] F. V. Jensen. Unconstrained influence diagrams. In *Proc. of the 18th International Conference on Uncertainty in Artificial Intelligence (UAI-02)*, pages 234–241. Morgan Kaufmann Publishers, 2002.

[10] S. L. Lauritzen and D. Nilsson. Representing and solving decision problems with limited information. *Management Science*, 47:1235–1251, 2001.

[11] S. L. Lauritzen and D. J. Spiegelhalter. Local computations with probabilities on graphical structures and their application to expert systems. *Journal of the Royal Statistical Society, Series B (Methodological)*, 50(2):157–224, 1988.

[12] R. Marinescu. A new approach to influence diagram evaluation. In *Proceedings of the 29th SGAI International Conference on Innovative Techniques and Applications of Artificial Intelligence*, Cambridge, UK, 2009.

[13] T. D. Nielsen and F. V. Jensen. Welldefined decision scenarios. In *Proceedings of the 15th Annual Conference on Uncertainty in Artificial Intelligence (UAI-99)*, pages 502–511, 1999.

[14] D. Nilsson and M. Hohle. Computing bounds on expected utilities for optimal policies based on limited information. Technical Report 94, Dina Research, 2001.

[15] S. M. Olmsted. *On representing and solving decision problems*. PhD thesis, Stanford University, Engineering-Economic Systems Department, 1983.

[16] J. Pearl. *Probabilistic Reasoning in Intelligent Systems: Networks of Plausible Inference*. Morgan Kaufmann Publishers, Inc., San Mateo, CA, 1988.

[17] R. Qi and D. Poole. A new method for influence diagram evaluation. *Computational Intelligence*, 11:498–528, 1995.

[18] R. D. Shachter. Evaluating influence diagrams. *Oper. Res.*, 34(6):871–882, 1986.

[19] P. Shenoy. Valuation based systems for Bayesian decision analysis. *Operations Research*, 40:463–484, 1992.

[20] C. Yuan and E. A. Hansen. Efficient computation of join-tree bounds for systematic MAP search. In *Proceedings of 21st International Joint Conference on Artificial Intelligence (IJCAI-09)*, pages 1982–1989, Pasadena, CA, 2009.

[21] N. L. Zhang. Probabilistic inference in influence diagrams. In *Computational Intelligence*, pages 514–522, 1998.